\documentclass[runningheads]{llncs}
\usepackage{graphicx}
\usepackage{threeparttable}
\usepackage{booktabs}
\usepackage{multirow}
\usepackage{tabularx}
\usepackage{amsmath}
\usepackage{amssymb}
\usepackage{amsfonts}
\usepackage{color}
\newcommand{\red}[1]{\color{black}#1\color{black}}
\begin{document}
\title{Deep Bayesian Active-learning-to-rank for Endoscopic Image Data}
%
\titlerunning{Deep Bayesian Active-learning-to-rank for Endoscopic Image Data}
%
\author{Takeaki Kadota\inst{1} \and
Hideaki Hayashi\inst{1} \and
Ryoma Bise\inst{1,2} \and
Kiyohito Tanaka\inst{3} \and
Seiichi Uchida\inst{1,2}
}
\authorrunning{T. Kadota et al.}
%
\institute{Kyushu University, Fukuoka, Japan \\
\email{takeaki.kadota@human.ait.kyushu-u.ac.jp} \and
National Institute of Informatics, Tokyo, Japan \and
Kyoto Second Red Cross Hospital, Kyoto, Japan}
\maketitle              
\begin{abstract}
Automatic image-based disease severity estimation generally uses discrete (i.e., quantized) severity labels. Annotating discrete labels is often difficult due to the images with ambiguous severity. An easier alternative is to use relative annotation, which compares the severity level between image pairs. By using a learning-to-rank framework with relative annotation, we can train a neural network that estimates rank scores that are relative to severity levels. However, the relative annotation for all possible pairs is prohibitive, and therefore, appropriate sample pair selection is mandatory. This paper proposes a \textit{deep Bayesian active-learning-to-rank}, which trains a Bayesian convolutional neural network while automatically selecting appropriate pairs for relative annotation. 
We confirmed the efficiency of the proposed method through experiments on endoscopic images of ulcerative colitis.  In addition, we confirmed that our method is useful even with the severe class imbalance because of its ability to select samples from minor classes automatically.

\end{abstract}
\section{Introduction\label{sec:intro}}
For image-based estimation of disease severity, it is common to prepare training samples with severity labels annotated by medical experts. Standard annotation (hereafter called \textit{absolute annotation}) assumes discretized severity labels. However, absolute annotation is often difficult to even for medical experts. This is because disease severity is not discrete in nature, and thus there are many ambiguous cases. For example, when annotating medical images with four discrete labels (0, 1, 2, and 3), they will frequently encounter medical images with a severity of 1.5. \par
\begin{figure}[t]
\centering
\includegraphics[scale=0.5]{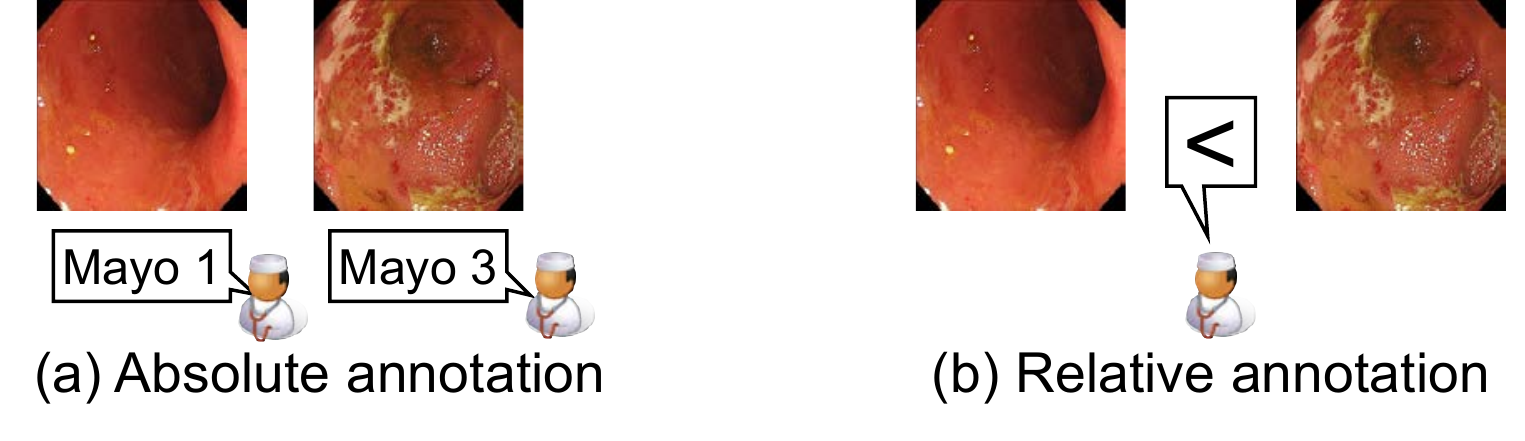}
\caption{Characteristics of absolute and relative annotations} \label{fig1}
\end{figure}

A promising alternative annotation approach is \textit{Relative annotation}, which compares two images for their severity and attaches a relative label that indicates the comparison result. Figure~\ref{fig1} shows the characteristics of absolute and relative annotations to endoscopic images of ulcerative colitis (UC). Relative annotation is far easier than absolute annotation and is expected to be stable even for real-valued targets~\cite{Parikh2011}. \par
Using image pairs with relative annotation, it is possible to train a \textit{ranking function} $f(x)$ that satisfies $f(x_i) > f(x_j)$ for a pair $(x_i, x_j)$ where $x_i$ shows a higher severity than $x_j$. By training $f(x)$ with many image pairs, it will give larger rank scores for severer images; in other words, the rank score $f(x)$ can be used as the severity level of $x$. As discussed later, the training process of $f(x)$ can be combined with representation learning to have a feature representation for better ranking. A convolutional neural network (CNN) is, therefore, a natural choice to realize $f(x)$.\par

A practically important issue in the above learning-to-rank framework with relative annotations is that we must select appropriate image pairs to annotate from all possible pairs. This is because we have $N(N-1)/2$ possible pairs for $N$ image samples. Even if individual relative annotations are easy, 
it is intractable to annotate all of them. In other words, careful selection of image pairs to be annotated is essential to fully demonstrate the efficiency of relative annotation. A naive selection strategy is a random selection; however, it easily overlooks minor but important samples, which often appear in medical applications. \par
In this paper, we propose a deep Bayesian active-learning-to-rank that fully maximizes the efficiency of relative annotation for efficient image-based severity estimation. The technical novelty of the proposed method is to introduce an \textit{active learning} technique into the learning-to-rank framework for selecting a less number of effective sample pairs. Active learning has been studied~\cite{Cohn1996} and generally takes the following steps. Firstly, a neural network is trained with a small amount of annotated samples. The trained network then suggests other samples to be annotated. These steps are repeated to have enough amount of the annotated training samples. \par
For suggesting important samples to be annotated, we employ an \textit{uncertainty} of the samples. If we find two samples with high uncertainty about their rank score, a new relative annotation between them will be useful to boost the ranking performance. For this purpose, we employ Bayesian CNN~\cite{Gal2017} to realize a ranking function because it can give an uncertainty of each sample. In summary, our deep Bayesian active-learning-to-rank uses Bayesian CNN for ranking with representation learning and uncertainty-based active learning. \par
As an experimental validation of the efficiency of the proposed method, we perform UC severity estimation using approximately 10,000 endoscopic images collected from the Kyoto Second Red Cross Hospital. Quantitative and qualitative evaluations clearly show the efficiency of the proposed framework. Especially, we reveal that the proposed method automatically selects minor but important samples.
\par

The main contributions of this paper are summarized as follows:
\begin{itemize}
\item We propose a deep Bayesian active-learning-to-rank that can rank the image samples according to their severity level. The proposed method can automatically select important image pairs for relative annotation, resulting in learning with a less number of relative annotations. 
\item Through UC severity estimation experiments, we demonstrated that the proposed method could suppress the number of relative annotations while keeping ranking performance.
\item We also experimentally confirmed that the proposed method shows robustness to class imbalance by its ability to select minor but important samples.
\end{itemize}

\section{Related work}
The application of deep learning to UC severity estimation has been studied~\cite{Stidham2019,Takenaka2020}. UC is a chronic inflammatory bowel disease with recurrent inflammation and ulcer recurrence in the colon. Accurate evaluation of treatment effects is important because the type and dosage of treatment medicines are adjusted in accordance with the condition of the patient with UC. Recently, Schwab et al.~\cite{Schwab2021} proposed an automatic UC severity estimation method for treatment effectiveness evaluation. Their method assumes a weakly supervised learning scenario because full annotation of all captured images is too costly. Since it uses absolute annotation, it will suffer from ambiguous labels, as noted above.
\par
Active learning has been widely applied to medical image analysis~\cite{Yang2017,Wen2018}.  Active learning selects the samples whose annotations are most effective for further learning from the unlabeled samples. Many uncertainty-based sampling techniques have been proposed to select highly uncertain samples as informative samples. Nair et al.~\cite{Nair2020} proposed a method using active learning for medical image analysis with uncertainty obtained from a Bayesian CNN. This method deals with an orthodox classification task with absolute annotation and thus does not assume any relative annotation and learning-to-rank.\par

In the natural language process (NLP) field, Wang et al.~\cite{Wang2021} recently introduced deep Bayesian active learning to learning-to-rank. Although their method sounds similar to ours, its aim and structure, as well as application, are totally different from ours. Specifically, it aims to rank sentences (called answers) to a fixed sentence (called a query) according to their relevance. For this aim, its network always takes two inputs (like $f(x_i, x_j)$) whereas ours takes a single input (like $f(x_i)$). From these differences, it is impossible to use it for the active annotation task and thus even to compare ours with it. 

\begin{figure}[t]
\centering
\includegraphics[scale=0.46]{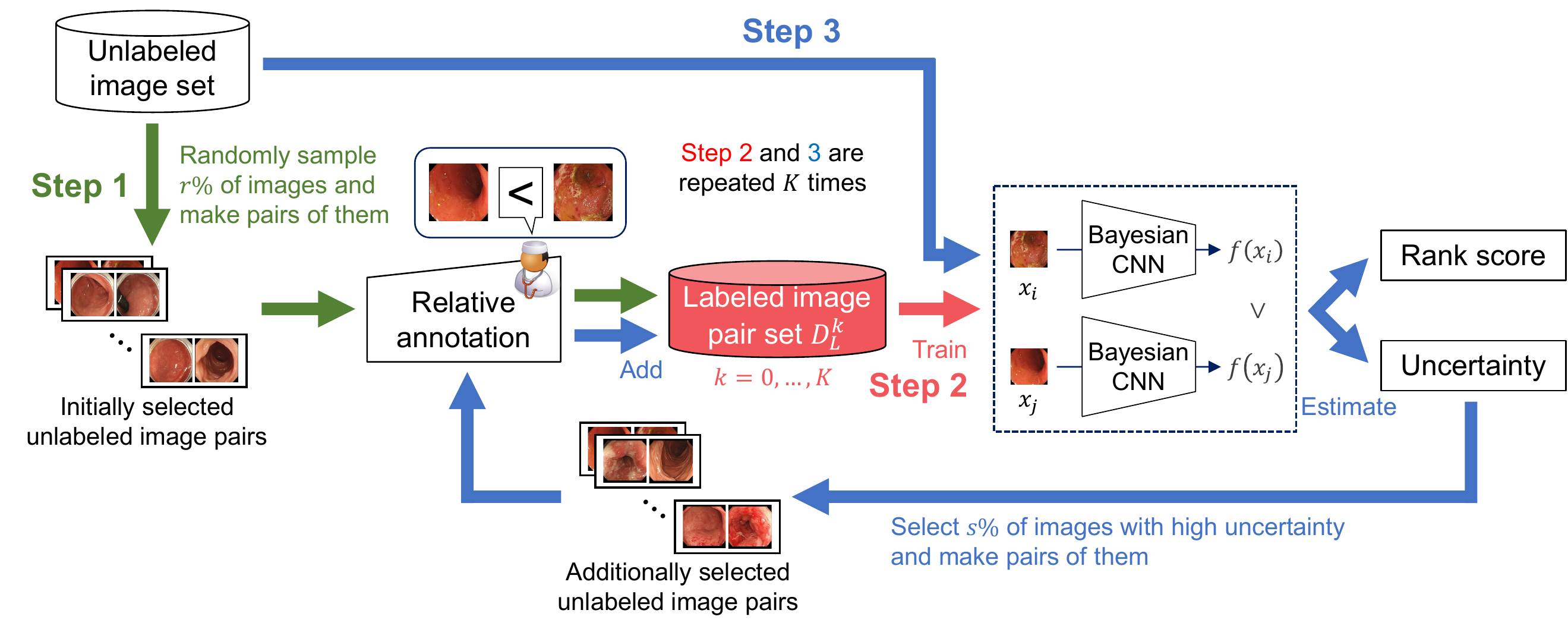}
\caption{Overall structure of proposed deep Bayesian active learning-to-rank for severity estimation of medical image data. } \label{fig2}
\end{figure}

\section{Deep Bayesian Active-learning-to-rank}
\subsection{Overview}
Figure~\ref{fig2} shows an overall structure of the proposed deep Bayesian active-learning-to-rank. The proposed method is organized in an active learning framework using Bayesian learning and learning-to-rank for severity estimation of medical image data; experts progressively add relative annotations to image pairs selected based on the uncertainty provided by a Bayesian CNN while training the Bayesian CNN based on the learning-to-rank algorithm. \par
The deep Bayesian active-learning-to-rank consists of three steps. In Step 1, for initial training, a small number of image pairs are randomly sampled from the unlabeled image set  and then annotated by medical experts. In Step 2, a Bayesian CNN is trained with the labeled image pairs for estimating rank scores and the uncertainties of individual training samples. In Step 3, images to be additionally labeled are selected on the basis of estimated uncertainties, having the medical experts annotate the additionally selected images via relative annotation. Steps 2 to 3 are repeated $K$ times to train the Bayesian CNN while progressively increasing the number of labeled image pairs. 
\vskip\baselineskip

\subsection{Details\label{sec:details}}
More details of each step are explained as follows. \par
\noindent
\textbf{Step 1: Preparing a small number of image pairs for initial training.} A set with a small number of annotated image pairs $\mathcal{D}^0_\mathrm{L}$ is prepared for initial training as follows. First, given an unlabeled image set comprising $N$ unlabeled images, we randomly sample $r\%$ of them (i.e., $R = rN/100$ image samples). 
Then, to form a set of $R$ pairs (instead of all possible $R(R-1)/2$ pairs), one of the $R-1$ samples is randomly paired for each of the $R$ samples\footnote{The strategy of making pairs is arbitrary. Here, we want to annotate all of the $R$ samples at least one time while avoiding $O(R^2)$ annotations and thus take this strategy.}.
Relative labels are attached to these image pairs by medical experts. For an image pair $(x_i,x_j)$, a relative label $C_{i,j}$ is defined as follows:
\begin{equation}
    C_{i,j}=
    \left\{ \begin{array}{ll}
    1, &  \mbox{if $x_i$ has a higher level than $x_j$,} \\
    0.5, & \mbox{else if $x_i$ and $x_j$ have the same level,}  \\
    0, & \mbox{otherwise.}  \\
    \end{array} \right.
\end{equation}
After this step, we have the annotated image pair set $\mathcal{D}^0_\mathrm{L} = \{(x_i, x_j, C_{i,j})\}$ and $|\mathcal{D}^0_\mathrm{L}|=R$.

\noindent
\textbf{Step 2: Training a Bayesian CNN.} A Bayesian CNN is trained as a ranking function with the labeled image pair set $\mathcal{D}^0_\mathrm{L}$. The CNN outputs a rank score that predicts the severity of the input image along with the uncertainty of the prediction. In training, we employ a probabilistic ranking cost function~\cite{Burges2005} to conduct learning-to-rank with a neural network and Monte Carlo (MC) dropout~\cite{Gal2016,Gal2017} for approximate Bayesian inference. \par
Let $f(\cdot)$ be a ranking function by a CNN with $L$ weighted layers. Given an image $x$, the CNN outputs a scalar value $f(x)$ as the rank score of $x$. We denote by $\mathbf{W}_l$ the $l$-th weight tensor of the CNN. For a minibatch $\mathcal{M}$ sampled from $\mathcal{D}^0_\mathrm{L}$, the Bayesian CNN is trained while conducting dropout with the loss function $\mathcal{L}_\mathcal{M}$ defined as follows:
\begin{equation}
    \mathcal{L}_\mathcal{M}=-\sum_{(i, j)\in\mathcal{I}_\mathcal{M}}\left\{C_{i,j}\log{P}_{i,j}+(1-C_{i,j})\log(1-{P}_{i,j})\right\} + \lambda \sum_{l=1}^L\|\mathbf{W}_l \|_F^2,
\label{eq:loss_function}
\end{equation}
where $\mathcal{I}_\mathcal{M}$ is a set of index pairs of the elements in $\mathcal{M}$, $P_{i,j} = \mathrm{sigmoid}(f(x_i)-f(x_j))$, $\lambda$ is a constant value for weight decay, and $\|\cdot\|_F$ represents a Frobenius norm. In Eq.~(\ref{eq:loss_function}), the first term is a probabilistic ranking cost function~\cite{Burges2005}, which allows the CNN to train rank scores, and the second term is a weight regularization term that can be derived from the Kullback–Leibler divergence between the approximate posterior and the posterior of the CNN weights~\cite{Gal2016}. The CNN is trained by minimizing the loss function $\mathcal{L}_\mathcal{M}$ for every minibatch while conducting dropout; a binary random variable that takes one with a probability of $p_\mathrm{dropout}$ is sampled for every unit in the CNN at each forward calculation, and the output of the unit is set to zero if the corresponding binary variable takes zero.


The rank score for an unlabeled image $x^*$ is predicted by averaging over the output of the trained Bayesian CNN with MC dropout as $y^*=\frac{1}{T}\sum_{t=1}^Tf(x^*;\omega_t)$, where $T$ is the number of MC dropout trials, $\omega_t$ is the $t$-th realization of a set of the CNN weights obtained by MC dropout, and $f(\cdot;\omega_t)$ is the output of $f(\cdot)$ given a weight set $\omega_t$. \par
The uncertainty of the prediction is defined as the variance of the posterior distribution of $y^*$. This uncertainty is used to select images to be annotated in the next step, playing an important role in achieving active learning. The variance of the posterior distribution, $\mathrm{Var}_{q(y^*|x^*)}[y^*]$, is approximately calculated using MC dropout as follows:
\begin{flalign}
\mathrm{Var}_{q(y^*|x^*)}[y^*] 
&=\mathbb{E}_{q(y^*|x^*)}[(y^*)^2] - \left(\mathbb{E}_{q(y^*|x^*)}[y^*]\right)^2 \nonumber \\
&\approx \frac{1}{T} \sum^T_{t=1}\left(f(x^*; \omega_t)\right)^2 - \left(\frac{1}{T}\sum^{T}_{t=1}f(x^*; \omega_{t})\right)^2 + \mathrm{const.},
\end{flalign}
where $q(y^*|x^*)$ is the posterior distribution estimated by the model. The constant term can be ignored because the absolute value of uncertainty is not required in the following step. \par

\vskip\baselineskip
\noindent
\textbf{Step 3: Uncertainty-based sample selection.} A new set of annotated image pairs is provided based on the estimated uncertainty and relative annotation. We estimate the rank scores and the related uncertainties for the unlabeled images, using the trained Bayesian CNN and select $s\%$ of images with high uncertainty. Image pairs are made by pairing the selected images in the same manner as Step 1, and medical experts attach relative annotations to the image pairs. The newly annotated image pairs are added to the current annotated image pair set $\mathcal{D}^0_\mathrm{L}$. The Bayesian CNN is retrained with the updated $\mathcal{D}^1_\mathrm{L}$. Steps 2 and 3 are repeated $K$ times while increasing the size of the annotated set $\mathcal{D}^k_\mathrm{L}$ ($k=0, \ldots, K$).

\section{Experiments and Results}
To evaluate the efficiency of our active-learning-to-rank, we conducted experiments on a task for estimating rank scores of ulcerative colitis (UC) severity. In the experiments, we quantitatively compared our method with baseline methods in terms of the accuracy of relative label estimation, that is, the correctness of identifying the higher severity image in a given pair of two endoscopic images. We also evaluated the relationship between the human-annotated absolute labels and the estimated rank scores.

In addition, we analyzed the reason why our method successfully improved the performance of relative label estimation. Especially, we analyze the relationship between the uncertainty and class prior and show that our uncertainty-based sample selection could mitigate the class imbalance problem and then finally improve the performance.


\begin{figure}[t]
\centering
\includegraphics[scale=0.54]{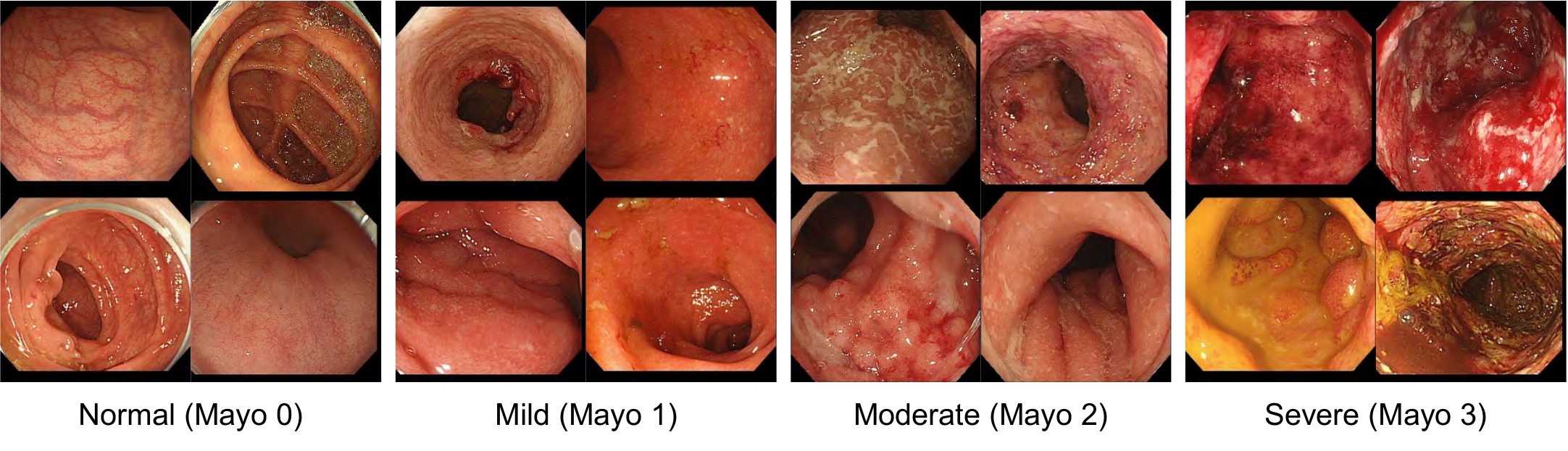}
\caption{Examples of endoscopic images of ulcerative colitis at each Mayo (severity).} \label{fig3}
\end{figure}

\subsection{Dataset}\label{sec:dataset}
In order to analyze the relationship between the absolute severity and the estimated severity scores, we used a dataset that has absolute severity labels (called Mayo score) for UC. A Mayo score was annotated for each image on a four-level scale (Mayo 0--3) by multiple medical experts carefully~\footnote{During this absolute annotation process, medical experts might encounter ambiguous cases due to the discretized severity, as noted in Section~\ref{sec:intro}. However, this ambiguity was minimized through careful observation by multiple experts. In other words, reliable absolute annotation is very costly, and this fact is the motivation of this work.}. Figure~\ref{fig3} shows examples of Mayo scores in the dataset. 
According to Schroeder et al.~\cite{Schroeder1987}, Mayo~0 is normal or endoscopic remission. Mayo~1 is a mild level that shows erythema, decreased vascular pattern, and mild friability. Mayo~2 is a moderate level that shows marked erythema, absent vascular pattern, friability, and erosions. Mayo~3 is a severe level with spontaneous bleeding and ulceration. The dataset consists of $10,265$ endoscopic images from 388 patients captured in the Kyoto Second Red Cross Hospital~\footnote{This study was approved by the Ethics Committee of the Kyoto Second Red Cross Hospital.}. It should be noted that the dataset has imbalanced class priors, 
which is a typical condition in medical image analysis. Specifically, it contains
6,678, 1,995, 1,395, and 197 samples for Mayo~0, 1, 2, and 3, respectively. \par
To evaluate relative label-based methods, we made a set of pairs of two images and gave relative labels for each pair based on the Mayo labels. For $N$ training samples, the number of possible pairs is too large ($O(N^2)$) to train the network with various settings; we, therefore, made a limited number of pairs for training data by random sampling. Specifically, we used $N$ pairs (instead of $O(N^2)$ pairs) by selecting one of the $N-1$ samples for each of the $N$ samples. (In Section~\ref{sec:details}, $R$ samples of the initial set are also selected from these pairs.)  We consider this setting reasonable since it used all the original samples, and we could conduct the experiments in a realistic running time. Also, note that this setting is typical for evaluating learning-to-rank \cite{2022Kadota,Xu2021,2019You}.\par

Five-fold cross-validation was conducted for all comparative methods. The data were divided into training (60\%), validation (20\%), and test (20\%) sets by patient-based sampling (that is, each data did not contain an image from the same patient). 
Note that the above pair image preparation was done after dividing the training, validation, and test sets for a fair evaluation. It indicates that each set did not contain the same image. As the performance metrics, we used the accuracy of estimated relative labels, which is defined as the ratio of the number of correctly estimated relative labels over all the pairs. 

\subsection{Implementation details}
We used DenseNet-169~\cite{Huang2017} as the backbone of the Bayesian CNN. The model was trained with dropout ($p_\mathrm{dropout}$ = 0.2) and weight decay ($\lambda$ = 1$\times$10$^{-4}$) in the convolutional and fully connected layers. We used Adam as the optimization algorithm and set the initial learning rate to 1$\times$10$^{-5}$. All image data were resized to $224 \times 224$ pixels and normalized between 0 and 255. \par
In all experiments, our method incrementally increased the training data during $K=6$ iterations by selecting effective samples ($s=5\%$ of training data) and adding them to the initial training data ($r=20\%$ of all training images). In total (after six iterations), the ratio of the labeled data used for training was $50\%$ ($r+sK=20+ 5\times 6$). The number of estimations for uncertainty estimation was set to $T = 30$.

\subsection{Baselines}
To demonstrate the effectiveness of our method, we compared our method with three baseline methods: 1) Baseline, which trained by randomly sampled $(r+sK=)~50\%$ pairs of training data, which indicates that the same number of pairs were used in the proposed method; 2) Baseline (all data), which uses all $N$ training pairs (i.e., 100\%), which indicates the number of training data was as twice as that of the proposed method; 3) Proposed w/o uncertainty-based sampling (UBS), which also incrementally increased the training data during $K$ iterations but the additional training data was selected by random sampling (without using uncertainty-based sampling); For a fair comparison, we used the same backbone (DenseNet-169 based Bayesian CNN) for all the methods. Given an input image, all methods used the mean of rank scores of $T=30$ times estimation as the rank score.\par
%
\subsection{Evaluation of relative severity estimation}\label{sec:eval}
As the accuracy evaluation, we measured the correctness of the estimated rank scores in a relative manner. Specifically, given a pair of images $(x_i, x_j)$ where $x_i$ shows a higher severity, the estimation is counted as ``correct'' if $f(x_i)>f(x_j)$.\par
For the test data, we prepared two types of pairs. \par
\noindent
\textbf{``Overall'':}\  The pairs were randomly made among all Mayo. Specifically, we selected images in test images so that the number of samples in each Mayo score was the same and then randomly made the pairs among the selected images.
Using this test dataset, we evaluated the overall performances of the comparative methods.
However, this test data may contain many easy pairs, that is, a pair of Mayo 0 (a normal image) and Mayo 3 (a high severity image), whose image features are very different, as shown in Fig.~\ref{fig3}. Thus it is easy to identify the relative label.\par\noindent
\textbf{``Neighboring'':}\  We prepared the neighboring pairs that contain the images of neighboring-level Mayo, such as Mayo~0--1, Mayo~1--2, and Mayo~2--3 pairs. It is important for clinical applications to compare the severity of difficult cases.  This estimation is more difficult since the severity gradually changes in neighboring Mayo, and these image features are similar. Using this test dataset, we evaluated the performance of methods in difficult cases.
\par

\newcolumntype{C}{>{\centering\arraybackslash}X}
\begin{table}[t]
  \centering
  \caption{Quantitative performance evaluation in terms of accuracy of estimated relative labels. \red{`*' denotes a statistically significant difference between the proposed method and each comparison method ($p<0.05$ in McNemar's test.).}}
  \label{tb:accracy}
    \scalebox{1}{
    \begin{tabularx}{\hsize}{lCCCCCC}
     \hline
     \multirow{2}{*}{Method}&\multirow{2}{*}{\begin{tabular}{c}Labeling\\ratio\end{tabular}}&\multirow{2}{*}{Overall}&\multicolumn{4}{c}{Neighboring}\\
     \cline{4-7}
     &&&0--1&1--2&2--3&Mean\\
     \hline \hline
     Baseline&50\%&0.861$^*$&0.827&0.837&0.628&0.763$^*$\\
     Baseline (all data)&100\%&0.875&{\bf0.855}&0.870&0.635&0.785$^*$\\
     Proposed w/o UBS&50\%&0.856$^*$&0.818&0.842&0.634&0.763$^*$\\
     Proposed&50\%&{\bf0.880}&0.787&{\bf0.871}&{\bf0.736}&{\bf0.797}\\
     \hline 
    \end{tabularx}
    }
\end{table}

Table~\ref{tb:accracy} shows the mean of the accuracy of estimated rank scores for each method in five-fold cross-validation.
A labeling ratio denotes the ratio of the number of labeled images that were used for training, and `*' indicates that there were significant differences at $p<0.05$ by multiple statistical comparisons using McNemar's tests.\par

In the results of ``Overall,'' the accuracy of Baseline and Proposed w/o UBS were comparable because these methods used the same number of training pairs. Baseline (all data) improved the accuracy compared to these two methods by using the larger size (twice) of training data.
Surprisingly, our method was better than Baseline (all data), although the training data of ours is half of that of Baseline (all data) with the same settings, that is, the network structure and the loss. As analyzed in the later Section~\ref{sec:analysis}, this difference comes from class imbalance. Since this dataset has a severe class imbalance, that is, the samples in Mayo 0 were 33 times of those in Mayo 3, it was difficult to learn the image features of highly severe images. In such severe imbalance cases, even if the number of training data decreases, appropriate sampling could mitigate the class imbalance and improve the accuracy.\par
In the results of ``Neighboring'' (difficult cases), the proposed method was also better than all the comparative methods in the mean accuracy of the neighboring pairs.
In particular, the proposed method improved the accuracy in the case ``Mayo 2--3'' over $10\%$. Since the accuracy of image comparisons at high severity is important for evaluating treatment effects, the proposed method is considered superior to the other methods in clinical practice.
In the case ``Mayo 0–1'', the accuracy of the proposed method was lower than that of the comparison methods because the number of labeled samples for Mayo~0 and 1 was reduced due to the mitigation of class imbalance.

\begin{figure}[t]
\centering
\includegraphics[scale=0.4]{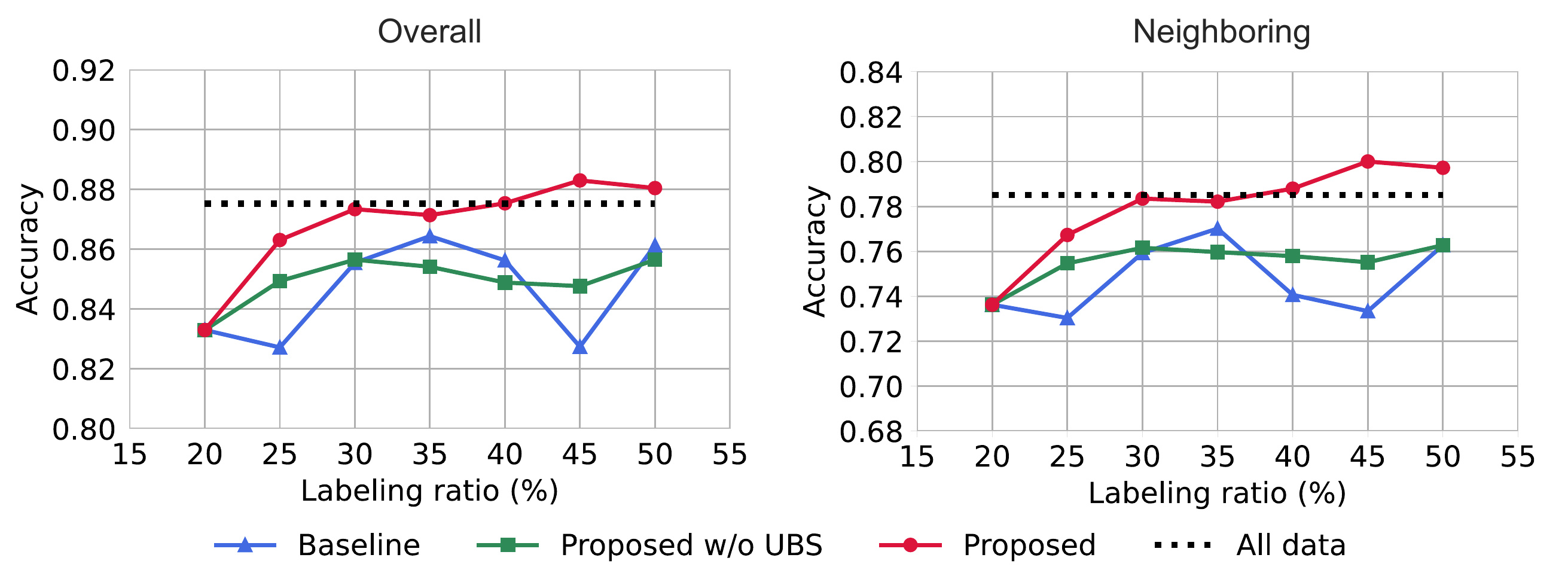}
\caption{Accuracy of estimated relative labels of Baseline (blue), Proposed w/o UBS (green), and Proposed (red) at each labeling ratio. The black dotted line indicates the results of Baseline (all data).}
\label{fig4}
\end{figure}

Figure~\ref{fig4} shows the changes of the accuracy at each iteration in both test datasets ``Overall'' and ``Neighboring.'' The horizontal axis indicates the labeling ratio, the vertical axis indicates the mean accuracy in cross-validation, and the black dot line indicates the results of Baseline (all data), which used the $100\%$ of the training data.  This result shows the effectiveness of our uncertainty-based active learning. Our method (red) increased the accuracy with the number of training data, and the improvement was larger than the other methods. In contrast, 
for Baseline (blue) and Proposed w/o UBS (green), it was not always true to increase the accuracy by increasing the training data. As a result, the improvements from the initial training data were limited for them.\par

\begin{figure}[t]
\centering
\includegraphics[scale=0.48]{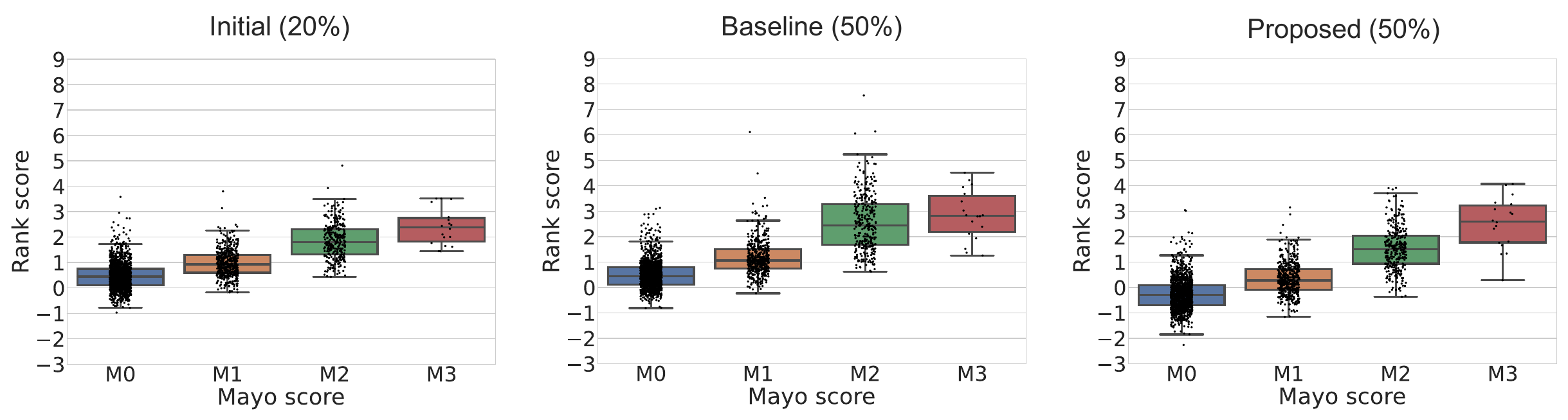}
\caption{ Box plots of estimated rank scores at each Mayo. \red{Initial was measured under the initial condition at the labeling ratio of $20\%$. Performance of Baseline and Proposed were measured at the labeling ratio of $50\%$. }If the distributions of each Mayo score have less overlap, the estimation can be considered reasonable.
}
\label{fig5}
\end{figure}

Figure~\ref{fig5} shows box plots of the estimated rank scores of three methods at each Mayo score.
Here, the vertical axis indicates the estimated rank score, and ``Initial'' indicates the method that trained the network using only the initial training data without iterations ($20\%$ of training data).
In this plot, if the distributions of each Mayo score have less overlap, the estimation can be considered reasonable.
In Initial and ``Baseline,'' the score distributions in Mayo 2 and 3 significantly overlapped. In contrast, our method improved the overlap distributions. This indicates that the estimated rank scores were more correlated to the Mayo scores.

\subsection{Relationship between uncertainty and class imbalance}\label{sec:analysis}
As described in Section~\ref{sec:eval}, we considered that improvement by our method is because our uncertainty-based sampling mitigated the class imbalance.
Therefore, we investigated the relationship between the uncertainty and the number of the Mayo labels of the sampled images from the training data during iterations.

\begin{figure}[t]
\centering
\includegraphics[scale=0.45]{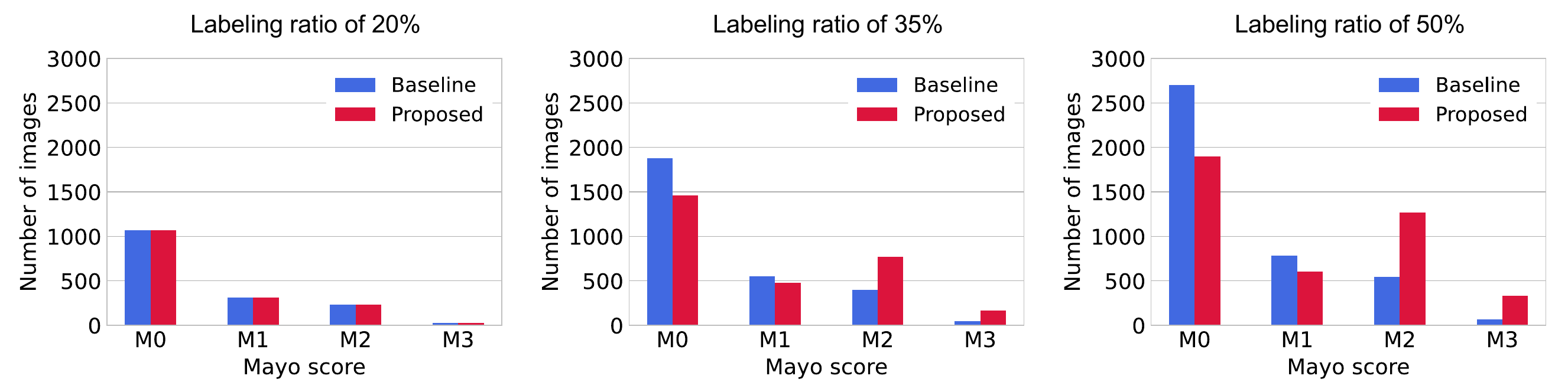}
\caption{Class proportion of the accumulated sampled images at iteration $K=0$ (labeling ratio is 20\%), 3 (35\%), and 6 (50\%). When the labeling ratio was 20\%, the class proportion was the same between Baseline and Proposed. The proposed method selected many samples of minor classes (Mayo 2 and 3) and mitigated the class imbalance problem.}
\label{fig6}
\end{figure}

\begin{figure}[t]
\centering
\includegraphics[scale=0.48]{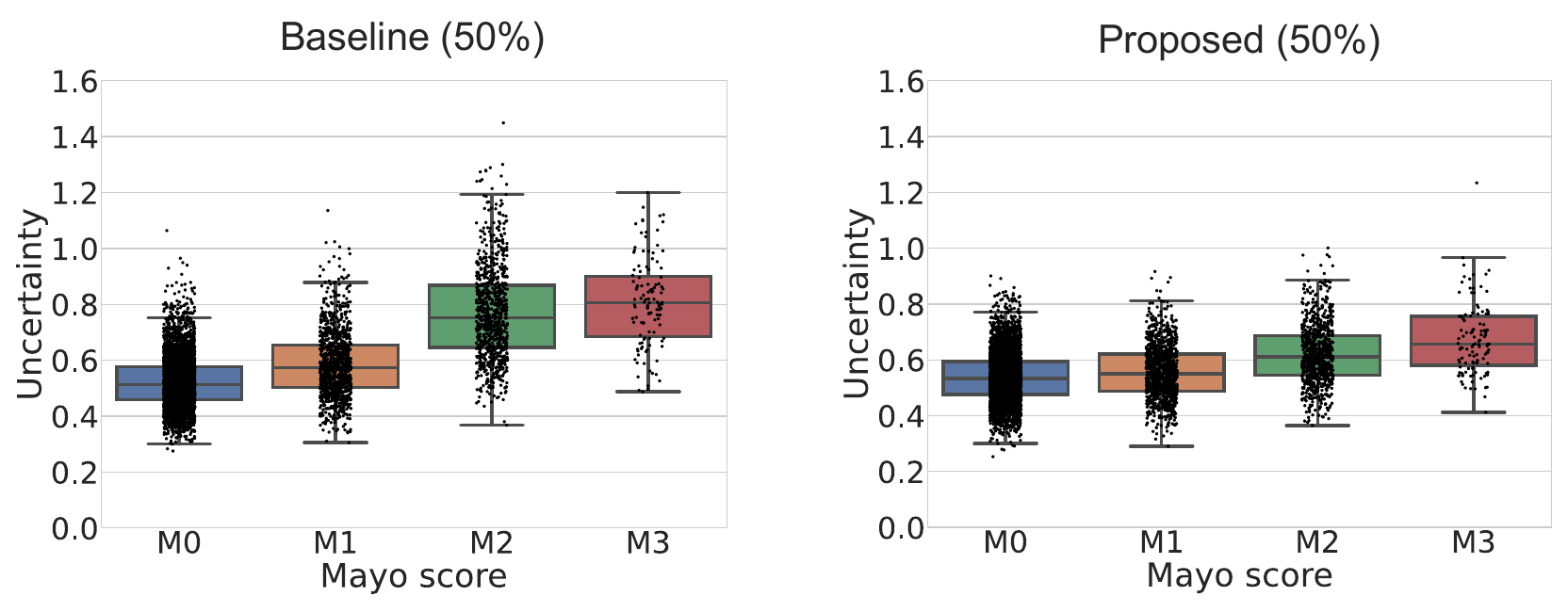}
\caption{Box plots of model uncertainty in each class. Performance of Baseline and Proposed were measured at the labeling ratio of $50\%$. In Baseline, the uncertainty of minor classes (Mayo 2 and 3) was higher than that of major classes. In Proposed, the uncertainty of Mayo 2 and 3 decreases since the class imbalance was mitigated.}
\label{fig7}
\end{figure}

Figure~\ref{fig6} shows the number of sampled images by uncertainty-based sampling at each class when the labeling ratio was 20\% ($k=0$), 35\% ($k=3$), and 50\% ($k=K=6$), where the vertical axis indicates the average numbers of five-fold cross-validation.
The initial training data ($k=0$) has a class imbalance in accordance with that in all the training data due to random sampling.
In 35\% and 50\%, the sampled images by Baseline have a similar class imbalance to that in the initial training data.
This class imbalance affected the performance improvements even though the number of training images increased, and thus the improvement was limited.
In contrast, our uncertainty-based sampling selected many samples of the minor Mayo level; that is, the samples of Mayo 2 and 3 increased with iteration; therefore, the class imbalance was gradually mitigated with iteration.
Consequently, the accuracy of relative label estimation was improved by our method despite the half size of training data compared to Baseline (all data).\par
Figure~\ref{fig7} shows the distributions of the model uncertainty of each class in training data.  In ``Baseline,'' the uncertainty of the minor classes (Mayo 2 and 3) was higher than the major classes (Mayo 0 and 1).
After six iterations of active learning (that is, the labeling ratio is 50\%), the class imbalance was mitigated, as shown in Fig.~\ref{fig6}, and the uncertainty of Mayo 2 and 3 was lower than those of ``Baseline.''
This indicates that the uncertainty is correlated to the class imbalance; the fewer samples of a class are, the higher uncertainty is.
Therefore, our uncertainty-based active learning, which selects the higher uncertainty samples as additional training data, can select many samples of the minor classes and consequently mitigate the class imbalance problem. This knowledge is useful for learning-to-rank tasks in medical image analysis since severe class imbalance problems often occur in medical images.

\begin{figure}[t]
\centering
\includegraphics[scale=0.47]{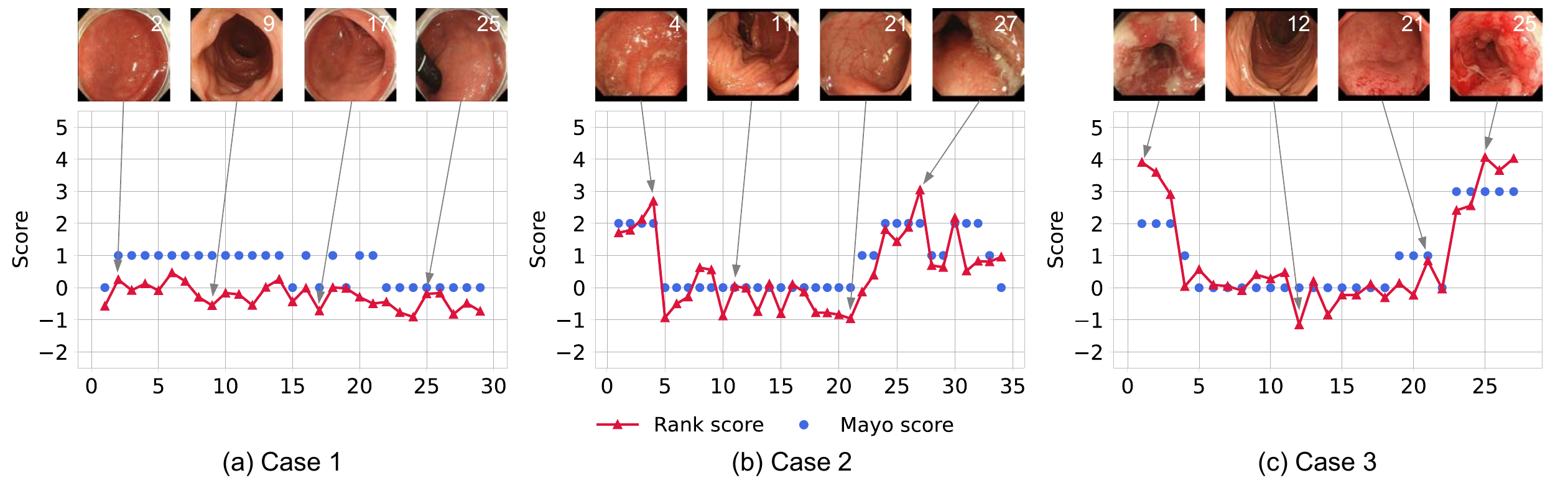}
\caption{Examples of changes in the rank score along with the capturing order in a sequence. The horizontal axis indicates the capturing order, and the vertical axis indicates the estimated rank score (red) and the Mayo score (blue).}
\label{fig8}
\end{figure}

\subsection{Application using the estimated rank score}
This section shows an example of the clinical applications of the proposed method.
In UC severity diagnosis, endoscopic images are acquired in sequence while the endoscope is moved through the colon. In clinical, a doctor checks all the images and finds the most severe images, and diagnoses the severity of the UC. To facilitate this  diagnosis process, it is useful to show the estimated severity score along with the capturing order.
Figure~\ref{fig8} shows three examples of the changes in the rank score (severity), where the horizontal axis indicates the capturing order in a sequence, and the vertical axis indicates the estimated rank score. These sequences were taken from different patients and were not used in the training data.
In these results, the estimated rank scores (red) were similar to the Mayo scores (blue) that were annotated by medical doctors.
Case 1 was a sequence of a patient with a mild disease level. The severity was low in the entire region.
Case 2 and 3 were sequences of patients with moderate and severe disease levels. In these examples, we can observe that the severe areas were biased in the order of the sequence; the severity was high at the beginning and the end of the sequence, but it was low in the middle of the sequence.
Using this graph, medical doctors can easily check if the disease is severe or not and find the severe areas and check them. In addition, it is also easy to show both cases before and after treatment and diagnose the recovery of the disease.

\section{Conclusion}
In this paper, we proposed a deep Bayesian active-learning-to-rank for efficient relative annotation. The proposed method actively determines effective sample pairs for additional relative annotations by estimating the uncertainty using a Bayesian CNN. We first evaluated the accuracy and the efficiency of the proposed method with an experiment about the correctness of the relative severity between a pair of images. The results indicate the usefulness of uncertainty-based active learning for selecting samples for better ranking. We also revealed that the proposed method selects minor but important samples and thus shows the robustness to  class imbalance. \par
The limitation of the proposed method is that it provides rank scores instead of Mayo scores that medical experts are familiar with. If an application needs Mayo score-like rank scores, an additional calibration process is necessary. The proposed method is very general and applicable to other severity-level estimation tasks --- this means more experiments on different image datasets are important for future tasks. 

\section*{Acknowledgments}
This work was supported by JSPS KAKENHI Grant Number JP20H04211 and AMED Grant Number JP20lk1010036h0002.

\bibliographystyle{splncs04}
\bibliography{refs}

\begin{thebibliography}{10}
\providecommand{\url}[1]{\texttt{#1}}
\providecommand{\urlprefix}{URL }
\providecommand{\doi}[1]{https://doi.org/#1}

\bibitem{Burges2005}
Burges, C., Shaked, T., Renshaw, E., Lazier, A., Deeds, M., Hamilton, N.,
  Hullender, G.: {Learning to rank using gradient descent}. In: Proceedings of
  the 22nd international conference on Machine learning (ICML). pp. 89--96
  (2005)

\bibitem{Cohn1996}
Cohn, D.A., Ghahramani, Z., Jordan, M.I.: {Active learning with statistical
  models}. Journal of Artificial Intelligence Research  \textbf{4},  129--145
  (1996)

\bibitem{Gal2016}
Gal, Y., Ghahramani, Z.: {Dropout as a Bayesian Approximation: Representing
  Model Uncertainty in Deep Learning}. In: Proceedings of the 33rd
  International Conference on Machine Learning (ICML). pp. 1050--1059 (2016)

\bibitem{Gal2017}
Gal, Y., Islam, R., Ghahramani, Z.: {Deep Bayesian active learning with image
  data}. In: Proceedings of the 34th International Conference on Machine
  Learning (ICML). pp. 1183--1192 (2017)

\bibitem{Huang2017}
Huang, G., Liu, Z., {Van Der Maaten}, L., Weinberger, K.Q.: {Densely connected
  convolutional networks}. In: 2017 IEEE Conference on Computer Vision and
  Pattern Recognition (CVPR). pp. 2261--2269 (2017)

\bibitem{2022Kadota}
Kadota, T., Abe, K., Bise, R., Kawamura, T., Sakiyama, N., Tanaka, K., Uchida,
  S.: {Automatic Estimation of Ulcerative Colitis Severity by Learning to Rank
  With Calibration}. IEEE Access  \textbf{10},  25688--25695 (2022)

\bibitem{Nair2020}
Nair, T., Precup, D., Arnold, D.L., Arbel, T.: {Exploring uncertainty measures
  in deep networks for Multiple sclerosis lesion detection and segmentation}.
  Medical Image Analysis  \textbf{59},  101557 (2020)

\bibitem{Parikh2011}
Parikh, D., Grauman, K.: {Relative Attributes}. In: Proceedings of the 2011
  International Conference on Computer Vision (ICCV). pp. 503--510 (2011)

\bibitem{Schroeder1987}
Schroeder, K.W., Tremaine, W.J., Ilstrup, D.M.: {Coated oral 5-aminosalicylic
  acid therapy for mildly to moderately active ulcerative colitis}. The New
  England Journal of Medicine  \textbf{317}(26),  1625--1629 (1987)

\bibitem{Schwab2021}
Schwab, E., Cula, G.O., Standish, K., Yip, S.S.F., Stojmirovic, A., Ghanem, L.,
  Chehoud, C.: {Automatic estimation of ulcerative colitis severity from
  endoscopy videos using ordinal multi-instance learning}. Computer Methods in
  Biomechanics and Biomedical Engineering: Imaging \& Visualization pp.~1--9
  (2021)

\bibitem{Stidham2019}
Stidham, R.W., Liu, W., Bishu, S., Rice, M.D., Higgins, P.D., Zhu, J.,
  Nallamothu, B.K., Waljee, A.K.: {Performance of a deep learning model vs
  human reviewers in grading endoscopic disease severity of patients with
  ulcerative colitis}. JAMA network open  \textbf{2}(5),  e193963 (2019)

\bibitem{Takenaka2020}
Takenaka, K., Ohtsuka, K., Fujii, T., Negi, M., Suzuki, K., Shimizu, H.,
  Oshima, S., Akiyama, S., Motobayashi, M., Nagahori, M., Saito, E., Matsuoka,
  K., Watanabe, M.: {Development and validation of a deep neural network for
  accurate evaluation of endoscopic images from patients with ulcerative
  colitis}. Gastroenterology  \textbf{158}(8),  2150--2157 (2020)

\bibitem{Wang2021}
Wang, Q., Wu, W., Qi, Y., Zhao, Y.: {Deep Bayesian Active Learning for Learning
  to Rank: A Case Study in Answer Selection}. IEEE Transactions on Knowledge
  and Data Engineering  (2021)

\bibitem{Wen2018}
Wen, S., Kurc, T., Hou, L., Saltz, J., Gupta, R., Batiste, R., Zhao, T.,
  Nguyen, V., Samaras, D., Zhu, W.: {Comparison of Different Classifiers with
  Active Learning to Support Quality Control in Nucleus Segmentation in
  Pathology Images}. AMIA Joint Summits on Translational Science Proceedings.
  AMIA Joint Summits on Translational Science  \textbf{2017},  227--236 (2018)

\bibitem{Xu2021}
Xu, Q., Yang, Z., Chen, Z., Jiang, Y., Cao, X., Yao, Y., Huang, Q.: {Deep
  Partial Rank Aggregation for Personalized Attributes}. In: Proceedings of the
  AAAI Conference on Artificial Intelligence. pp. 678--688 (2021)

\bibitem{Yang2017}
Yang, L., Zhang, Y., Chen, J., Zhang, S., Chen, D.Z.: {Suggestive Annotation: A
  Deep Active Learning Framework for Biomedical Image Segmentation}. In:
  Medical Image Computing and Computer Assisted Intervention (MICCAI). pp.
  399--407 (2017)

\bibitem{2019You}
You, Y., Lu, C., Wang, W., Tang, C.K.: {Relative CNN-RNN: Learning Relative
  Atmospheric Visibility From Images}. IEEE Transactions on Image Processing
  \textbf{28}(1),  45--55 (2019)

\end{thebibliography}
%




\end{document}